\documentclass{article}

\usepackage[preprint, nonatbib]{neurips_2024}

\usepackage[utf8]{inputenc} 
\usepackage[T1]{fontenc}    
\usepackage{hyperref}       
\usepackage{url}            
\usepackage{booktabs}       
\usepackage{amsfonts}       
\usepackage{nicefrac}       
\usepackage{microtype}      
\usepackage{xcolor}         

\usepackage{amsmath}
\usepackage{amssymb}
\usepackage{graphicx}
\usepackage[numbers]{natbib}
\usepackage{algorithm2e}
\usepackage{siunitx} 
\usepackage{tikz}
\usetikzlibrary{shapes.geometric, arrows, positioning, fit, calc}

\title{Cooperative Perception: A Resource-Efficient Framework for Multi-Drone 3D Scene Reconstruction Using Federated Diffusion and NeRF}

\author{%
  Massoud Pourmandi, PhD \\
  Department of Electrical and Electronics Engineering \\
  Boğaziçi University \\
  \texttt{massoud.pourmandi@std.bogazici.edu.tr} \\
}

\begin{document}

\maketitle

\begin{abstract}
The proposal introduces an innovative drone swarm perception system that aims to solve problems related to computational limitations and low-bandwidth communication, and real-time scene reconstruction. The framework enables efficient multi-agent 3D/4D scene synthesis through federated learning of shared diffusion model and YOLOv12 lightweight semantic extraction and local NeRF updates while maintaining privacy and scalability.  The framework redesigns generative diffusion models for joint scene reconstruction \cite{chen2025fedddpm, zhang2025collaborative}, and improves cooperative scene understanding \cite{chen2023cooperative}, while adding semantic-aware compression protocols \cite{qin2023semantic}. The approach can be validated through simulations and potential real-world deployment on drone testbeds, positioning it as a disruptive advancement in multi-agent AI for autonomous systems.
\end{abstract}

\section{Introduction}
The quest for Level 4/5 autonomy in intelligent systems operating in dynamic unstructured environments has exposed basic limitations of single agents \cite{badue2021self, huang2022multimodal, zhang2023deep}. The onboard sensing capabilities have made significant progress with high-resolution cameras, LiDAR, and RADAR, supported by large-scale datasets like nuScenes and Waymo \cite{caesar2020nuscenes, sun2020scalability, lu2024v2xreal}, but a single agent's perception is inherently limited by its physical viewpoint. This results in occlusions, sensor noise, and a limited field of view \cite{varghese2021survey, yoon2023survey, wang2024survey_v2v}, creating a critical safety bottleneck that prevents robust decision-making in complex scenarios such as dense urban traffic, search and rescue operations, or precision agriculture \cite{li2024multi, arnold2019survey}. Large-scale initiatives attempt to solve this problem by developing cooperative perception which enables a network of agents such as a swarm of unmanned aerial vehicles (UAVs) to share and fuse their local sensor data to create a unified comprehensive model of the world that is much more complete than what any single agent could achieve alone \cite{xu2023v2x, xu2022opencood}.

The motivation for cooperative perception is clear: by aggregating information from diverse viewpoints, a multi-agent system can eliminate occlusions, increase perceptual range, and enhance robustness to individual sensor failures, thereby enabling complex, coordinated tasks \cite{li2021learning, qin2021hoper, xu2022cobevt, chen2023cooperative}. The initial solutions to this challenge involved either \textit{late fusion} where agents exchanged high-level object detections or tracks \cite{li2022deepfusion, xu2022cobevt} or \textit{early fusion} where raw sensor data was sent to a central node for processing \cite{chen2019cooper}. The bandwidth efficiency of late fusion comes at the cost of discarding important low-level sensor data which is vital for understanding scene geometry and context \cite{yu2022dair}. Early fusion maintains all information but results in excessive communication costs which makes it unsuitable for real-time applications in UAV swarms operating in bandwidth-limited environments \cite{li2024multi, wang2020v2vnet, haydari2024survey}. 

The search for intermediate fusion has intensified because researchers aim to discover a "sweet spot" through agent compression of mid-level feature representations \cite{xu2022opv2v, li2023v2x-vit, huang2023bridging, hao2025lif}. The methods V2VNet \cite{wang2020v2vnet}, F-Cooper \cite{li2021learning_fcooper}, and Where2comm \cite{hu2022where2comm} have achieved substantial performance improvements in collaborative 3D object detection tasks. The methods depend on manual communication protocols or attention mechanisms to determine feature sharing but the probable large size of shared features creates scalability issues \cite{hu2023when2com, li2023learning, liu2020who2com, yazgan2024survey}. The approaches primarily concentrate on discriminative tasks while lacking built-in capabilities to create persistent queryable 3D environmental representations.

Concurrently, the field of 3D scene representation has been revolutionized by Neural Radiance Fields (NeRFs) \cite{mildenhall2020nerf} and, more recently, by explicit scene representations like 3D Gaussian Splatting \cite{kerbl20233d}, which offer unprecedented rendering quality and speed. The development of more efficient architectures like Instant-NGP \cite{muller2022instant}, Plenoxels \cite{fridovich2022plenoxels}, and streamable Gaussians \cite{wei2025lightweight} has made real-time 3D reconstruction feasible, inspiring a wave of research into their use for robotics tasks like mapping and SLAM \cite{sucar2021imap, zhu2022nice, chen2022sem2nerf, yan2023coslam}. This has naturally led to the concept of distributed or multi-agent scene reconstruction \cite{zhao2024distributed, li2022read, huang2023snerf, xu2023nerfdet}. Current paradigms, however, often require agents to share either raw images \cite{deng2022depth}, dense feature grids \cite{reiser2023merf}, or frequent, high-dimensional gradient updates \cite{zhu2023urban, xie2023cnerf}, reintroducing the very communication and computation bottlenecks that cooperative perception seeks to avoid. While federated learning has been proposed to train these models in a privacy-preserving manner \cite{siddiqui2022fednerf, tasneem2024decentnerf, suzuki2024fed3dgs}, these methods typically focus on offline training and are not yet suited for real-time, dynamic scene reconstruction in the field.

This reveals a critical, unaddressed gap at the intersection of cooperative perception and generative 3D scene modeling. The central challenge is a trilemma of competing constraints:
\begin{enumerate}
    \item \textbf{Communication Bottleneck:} How can agents share information to build a coherent global scene without overwhelming low-bandwidth wireless channels \cite{haydari2024survey}?
    \item \textbf{Computational Burden:} How can complex generative models for 3D reconstruction be deployed on resource-constrained platforms like drones, which have limited processing power and battery life \cite{li2024multi}?
    \item \textbf{Privacy and Scalability:} How can a swarm build a shared understanding without centralizing raw data, thus preserving privacy and ensuring the system scales to a large number of agents \cite{konevcny2016federated}?
\end{enumerate}

The proposed framework tackles this trilemma through a new approach that transforms how agents share information. Agents should exchange only condensed semantic information according to this proposed paradigm because this concept aligns with current developments in semantic communication \cite{qin2023semantic, shannon1948mathematical, lan2024generative, zhang2023task}. The main breakthrough occurs through the deployment of a 3D-aware generative prior diffusion model \cite{ho2020ddpm, rombach2022high} which receives collaborative training through federated learning \cite{chen2025fedddpm, mcmahan2017communication, zhang2023feddiff, wang2025fedphd} to hallucinate photorealistic 2D images of unobserved areas from sparse semantic and pose prompts. The hallucinated views serve as training data to develop or enhance local NeRF or Gaussian Splatting models \cite{yi2023gaussiandreamer} because they maintain geometric consistency through precise pose conditioning \cite{zhang2023adding, poole2022dreamfusion} from occluded or distant collaborators.

The proposed solution disrupts the existing problems through its innovative approach. The proposed framework is expected to decrease communication overhead compared to feature-sharing methods according to \cite{li2021learning_fcooper}. The system distributes the heavy generative work to the requesting agent while source agents execute lightweight semantic extraction using YOLOv12 \cite{tian2025yolov12} or CLIP \cite{radford2021learning} models. The system maintains privacy and scalability through federated learning, which enables the development of a shared powerful diffusion model without requiring private sensor data exchange \cite{konevcny2016federated, chen2024phoenix}. The framework combines semantic communication principles \cite{qin2023semantic, shannon1948mathematical} with federated generative modeling \cite{gossmann2022federated} and contemporary neural scene representation \cite{kerbl20233d, wu20244d} to create a new generation of lean intelligent scalable multi-agent autonomous systems.

\section{Background}
\subsection{Neural Radiance Fields (NeRF)}
Neural Radiance Fields (NeRF) represent a continuous 3D scene as a 5D function, implemented by a multilayer perceptron (MLP), that maps a 3D coordinate $\mathbf{x}=(x,y,z)$ and a 2D viewing direction $\mathbf{d}=(\theta, \phi)$ to a volume density $\sigma$ and an emitted color $\mathbf{c}$ \citep{mildenhall2020nerf}.
\[
F_\Theta: (\mathbf{x}, \mathbf{d}) \rightarrow (\mathbf{c}, \sigma)
\]
Novel views are synthesized using principles of classical volume rendering. The color $C(\mathbf{r})$ of a camera ray $\mathbf{r}(t) = \mathbf{o} + t\mathbf{d}$ is computed by integrating the color and density along the ray:
\[
C(\mathbf{r}) = \int_{t_n}^{t_f} T(t) \sigma(\mathbf{r}(t)) \mathbf{c}(\mathbf{r}(t), \mathbf{d}) dt, \quad \text{where} \quad T(t) = \exp\left(-\int_{t_n}^{t} \sigma(\mathbf{r}(s)) ds\right).
\]
Here, $T(t)$ is the accumulated transmittance, representing the probability that the ray travels from the near bound $t_n$ to $t$ without being occluded. The MLP is optimized by minimizing the photometric error between rendered and ground truth pixel colors.

\subsection{Diffusion Models}
Diffusion models are a class of generative models that learn to reverse a gradual noising process \citep{ho2020ddpm}. They consist of two processes:
\begin{enumerate}
    \item \textbf{Forward Process (Diffusion):} A fixed Markov chain gradually adds Gaussian noise to data $\mathbf{x}_0$ over $T$ timesteps, according to a variance schedule $\{\beta_t\}_{t=1}^T$.
    \[
    q(\mathbf{x}_t | \mathbf{x}_{t-1}) = \mathcal{N}(\mathbf{x}_t; \sqrt{1 - \beta_t} \mathbf{x}_{t-1}, \beta_t \mathbf{I})
    \]
    This allows sampling $\mathbf{x}_t$ at any timestep directly from $\mathbf{x}_0$: $\mathbf{x}_t = \sqrt{\bar{\alpha}_t}\mathbf{x}_0 + \sqrt{1 - \bar{\alpha}_t}\boldsymbol{\epsilon}$, where $\alpha_t = 1 - \beta_t$, $\bar{\alpha}_t = \prod_{s=1}^t \alpha_s$, and $\boldsymbol{\epsilon} \sim \mathcal{N}(\mathbf{0}, \mathbf{I})$.

    \item \textbf{Reverse Process (Denoising):} A neural network $\epsilon_\theta$ is trained to predict the added noise $\boldsymbol{\epsilon}$ from the noisy image $\mathbf{x}_t$ at timestep $t$. The model learns the reverse process $p_\theta(\mathbf{x}_{t-1} | \mathbf{x}_t)$ to reconstruct the data by iteratively denoising, starting from pure noise $\mathbf{x}_T \sim \mathcal{N}(\mathbf{0}, \mathbf{I})$. The training objective is typically a simplified mean squared error loss:
    \[
    \mathcal{L}_{\text{simple}} = \mathbb{E}_{t, \mathbf{x}_0, \boldsymbol{\epsilon}} \left\| \boldsymbol{\epsilon} - \epsilon_\theta(\sqrt{\bar{\alpha}_t}\mathbf{x}_0 + \sqrt{1 - \bar{\alpha}_t}\boldsymbol{\epsilon}, t) \right\|^2.
    \]
\end{enumerate}

\subsection{Federated Learning (FL)}
Federated Learning (FL) is a distributed machine learning paradigm that enables model training on decentralized data located on devices like drones without exchanging the data itself \citep{mcmahan2017communication}. The most common algorithm is Federated Averaging (FedAvg). The process involves:
\begin{enumerate}
    \item A central server initializes a global model and distributes it to a set of clients.
    \item Each client trains the model on its local data for a few epochs.
    \item Clients send their updated model weights (or gradients) back to the server.
    \item The server aggregates the client updates to produce a new global model, typically by weighted averaging:
    \[
    \theta_{t+1} = \sum_{k=1}^{K} \frac{n_k}{n} \theta_{t+1}^k,
    \]
    where $\theta_{t+1}$ is the new global model, $K$ is the number of clients, $n_k$ is the number of data points on client $k$, $n$ is the total number of data points, and $\theta_{t+1}^k$ is the model update from client $k$.
\end{enumerate}
This iterative process continues until the global model converges, ensuring data privacy and reducing communication overhead.

\subsection{Multiterminal Coding}
Multiterminal coding, also known as Distributed Source Coding (DSC), provides the theoretical foundation for compressing data from multiple correlated sources whose encoders operate independently but whose decoders can operate jointly \citep{slepian1973noiseless}. This is the exact scenario in cooperative perception, where drones (encoders) observe a correlated scene but a target drone (or a downstream fusion center) acts as a joint decoder.

\paragraph{The Slepian-Wolf Theorem (Lossless Coding)}
This theorem defines the achievable rate region for the lossless compression of two correlated sources, $X$ and $Y$. While a naive approach would require a total rate of $H(X) + H(Y)$, the joint entropy $H(X,Y)$ is lower due to the correlation. The Slepian-Wolf theorem states that the sources can be encoded separately and recovered perfectly at a joint decoder, as long as the individual rates $R_X$ and $R_Y$ satisfy:
\begin{align*}
R_X &\geq H(X|Y) \\
R_Y &\geq H(Y|X) \\
R_X + R_Y &\geq H(X,Y)
\end{align*}
Here, $H(X|Y)$ is the conditional entropy. This is a profound result: one agent can compress its data as if it already knew the other agent's observation, with the correlation being exploited entirely at the decoder.

\paragraph{The Wyner-Ziv Theorem (Lossy Coding with Side Information)}
This extends the concept to lossy compression, where one agent's data ($Y$) is available as "side information" at the decoder. The Wyner-Ziv theorem states that the source agent ($X$) can achieve the same rate-distortion performance as if its encoder also had access to $Y$ \citep{wyner1976rate}. This is highly relevant for our framework, where a target drone's local view can serve as side information to decompress messages from other drones. Recent extensions incorporate AI to adaptively model the side information, such as using shared world models to estimate correlations and further reduce transmission rates \citep{ahmadipour2023hybrid}.

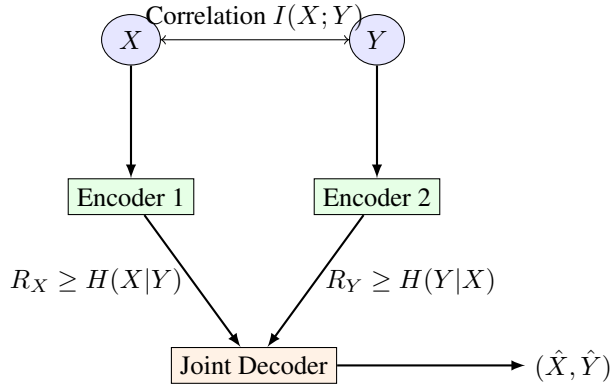
\begin{figure}[h!]
\centering
\begin{tikzpicture}[
node distance=1.5cm and 2.5cm,
source/.style={draw, ellipse, fill=blue!10},
encoder/.style={draw, rectangle, fill=green!10},
decoder/.style={draw, rectangle, fill=orange!10, minimum width=2cm},
arrow/.style={-latex, thick}
]
\node[source] (X) {$X$};
\node[source, right=of X] (Y) {$Y$};
\draw[<->] (X) -- (Y) node[midway, above] {Correlation $I(X;Y)$};
\node[encoder] (EncX) [below=of X] {Encoder 1};
\node[encoder] (EncY) [below=of Y] {Encoder 2};
\draw[arrow] (X) -- (EncX);
\draw[arrow] (Y) -- (EncY);
\node[decoder] (Dec) [below=2cm of $(EncX)!0.5!(EncY)$] {Joint Decoder};
\draw[arrow] (EncX) -- (Dec) node[midway, left] {$R_X \ge H(X|Y)$};
\draw[arrow] (EncY) -- (Dec) node[midway, right] {$R_Y \ge H(Y|X)$};
\node (XY_hat) [right=of Dec] {$(\hat{X}, \hat{Y})$};
\draw[arrow] (Dec) -- (XY_hat);
\end{tikzpicture}
\caption{The Slepian-Wolf multiterminal coding framework. Correlated sources $X$ and $Y$ are encoded independently but decoded jointly, achieving a total rate approaching the joint entropy $H(X,Y)$.}
\label{fig:slepian_wolf}
\end{figure}

\subsection{Semantic Compression}
Semantic compression redefines the goal of compression from faithfully reconstructing the original data (syntactic fidelity) to preserving the information necessary for a specific downstream task (semantic fidelity). Instead of minimizing a metric like Mean Squared Error (MSE), $d(x, \hat{x}) = ||x - \hat{x}||^2$, it seeks to minimize a \textbf{semantic distortion}, $d_S(x, \hat{x})$, which measures how the compression affects task performance \citep{blau2019rethinking}.

Let $f(x)$ be the output of a perception task (e.g., object detection) on the original data $x$. The semantic distortion can be defined as the difference in task performance on the reconstructed data $\hat{x}$:
\[
d_S(x, \hat{x}) = \mathcal{L}_{\text{task}}(f(x), f(\hat{x}))
\]
where $\mathcal{L}_{\text{task}}$ is the task's loss function. The goal is to minimize the transmission rate $R$ subject to the constraint that the expected semantic distortion remains below a tolerable threshold. Recent methods enhance this by using "importance maps" to dynamically weight distortion by semantic relevance, assigning higher priority to critical regions like moving objects \citep{qin2023semantic}.

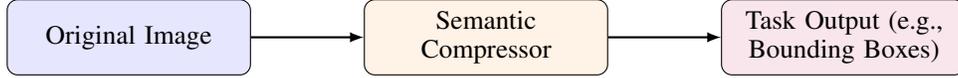
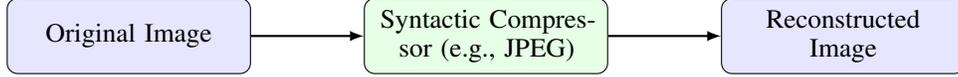
\begin{figure}[h!]
\centering
\begin{tikzpicture}[
node distance=1.5cm,
block/.style={draw, rectangle, rounded corners, minimum height=1cm, text width=3cm, align=center},
arrow/.style={-latex, thick}
]
\node (img1) [block, fill=blue!10] {Original Image};
\node (comp1) [block, fill=green!10, right=of img1] {Syntactic Compressor (e.g., JPEG)};
\node (rec1) [block, fill=blue!10, right=of comp1] {Reconstructed Image};
\node (mse) [below=of comp1, yshift=-0.5cm] {Goal: Minimize $||X - \hat{X}||^2$};
\draw[arrow] (img1) -- (comp1);
\draw[arrow] (comp1) -- (rec1);
\node (img2) [block, fill=blue!10, below=of img1, yshift=-2.0cm] {Original Image};
\node (comp2) [block, fill=orange!10, right=of img2] {Semantic Compressor};
\node (rec2) [block, fill=purple!10, right=of comp2] {Task Output (e.g., Bounding Boxes)};
\node (task_loss) [below=of comp2, yshift=-0.5cm] {Goal: Minimize $\mathcal{L}_{task}(f(X), f(\hat{X}))$};
\draw[arrow] (img2) -- (comp2);
\draw[arrow] (comp2) -- (rec2);
\end{tikzpicture}
\caption{Comparison of syntactic vs. semantic compression pipelines. Syntactic compression aims to reconstruct the input data, while semantic compression aims to preserve only the information needed for a specific downstream task.}
\label{fig:semantic_comp}
\end{figure}

\section{Proposed Framework}
\subsection{System Setup}

The system includes $N$ drones that operate in separate zones. The drones use local Neural Radiance Fields (NeRF) for 3D scene representation through efficient variants such as Instant-NGP \citep{muller2022instant} and MobileNeRF to meet drone hardware requirements.

The shared generative diffusion model based on Stable Diffusion \citep{rombach2022high} learns 3D-consistent generation through explicit camera pose and semantic prompt conditioning. The core component of our cooperative strategy uses federated learning (FL) to train this model. The drones process their local data, consisting of images with semantic information and pose data, which they send to a target aggregator to develop a global prior for scene hallucination that represents multiple environments.

Resource constraints are addressed by sharing only semantics (e.g., text tokens or embeddings, of \SI{}{\kilo\byte} size) and poses (6-DoF matrices). This ensures low bandwidth ($<$\SI{1}{\mega\byte} per exchange) and latency (milliseconds over Wi-Fi/5G). Source drones extract semantics lightweightly using pre-trained models like CLIP \citep{radford2021learning}, while the target drone handles diffusion inference (quantized for edge devices) and NeRF updates.

\subsection{Cooperative Mechanism}
When Drone 1 (target) incorporates zones from Drones 2 and 3:

1. \textbf{Request Phase:} Drone 1 broadcasts a request specifying viewpoints or zones.

2. \textbf{Data Sharing Phase:} Source drones covering the requested zones respond with semantic info (e.g., "winding path, exposed tree roots") extracted via YOLOv12 \citep{tian2025yolov12} and reference poses in a shared coordinate frame (aligned via GPS/IMU).

3. \textbf{Hallucination Phase:} Drone 1 conditions its local diffusion model on received semantics (text prompts) and poses, using adapters like ControlNet \citep{zhang2023adding} to precisely control the viewpoint of the output. This pose conditioning is the critical step that ensures the generated 2D images are multi-view consistent and suitable for reconstructing a coherent 3D scene with NeRF. The generation process can be abstracted as a function call to the generative model:
    \[
    I_p = \text{GenerateImage}(\theta, c_p)
    \]
    where the model with parameters $\theta$ takes in conditioning information $c_p$, which contains both the fused semantics and the specific pose $p$, to produce a corresponding image $I_p$.

4. \textbf{NeRF Construction/Update Phase:} Hallucinated images and poses are used to incrementally train the local NeRF over a batch of rays $\mathcal{R}$:
    \[
    \mathcal{L} = \sum_{\mathbf{r} \in \mathcal{R}} \left\| \hat{C}(\mathbf{r}) - C(\mathbf{r}) \right\|^2,
    \]
    where $\hat{C}(\mathbf{r})$ is the rendered color for a ray $\mathbf{r}$ and $C(\mathbf{r})$ is the ground truth color from the hallucinated images.

5. \textbf{Feedback and Iteration:} Optional validation and refinements, with ongoing FL to improve the model.

\RestyleAlgo{ruled}
\SetKwComment{Comment}{/* }{ */}
\begin{algorithm}[H]
\caption{Cooperative Perception Algorithm}
\KwData{Target drone $D_1$, All source drones $\{D_2, \dots, D_N\}$, Shared generative model $\theta$}
\KwResult{Updated NeRF for $D_1$}
$D_1$ broadcasts request for information on specific zones\;
Let $\mathcal{D}_{resp} \subseteq \{D_2, \dots, D_N\}$ be the set of responding drones covering the requested zones\;
\For{each source $D_i \in \mathcal{D}_{resp}$}{
    Extract semantics $S_i$ using YOLOv12\;
    Compute reference poses $P_i$ in the global coordinate frame\;
    Send $(S_i, P_i)$ to $D_1$\;
}
$S_{fused} \gets \text{Fuse}(\{S_i\}_{i \in \mathcal{D}_{resp}})$ \Comment{Fuse semantic information}
$P_{all} \gets \bigcup_{i \in \mathcal{D}_{resp}} P_i$ \Comment{Aggregate all poses}
\For{each pose $p \in P_{all}$}{
    $c_p \gets (S_{fused}, p)$ \Comment{Combine semantics and pose for conditioning}
    Hallucinate image $I_p = \text{GenerateImage}(\theta, c_p)$\;
}
Update NeRF: Train on the set of hallucinated images and poses $\{(I_p, p)\}$ incrementally\;
\If{validation of the reconstructed scene fails}{
    Request refinements from source drones\;
}
Contribute to FL: Periodically share local gradients to update the global model $\theta$\;
\end{algorithm}

\section{Technical Details}
\subsection{Semantic Extraction with YOLOv12}
YOLOv12 \citep{tian2025yolov12}, released February 18, 2025, incorporates attention mechanisms (e.g., Area Attention) for real-time object detection and segmentation. It achieves low latency ($\sim$\SI{10}{\milli\second} on edge GPUs) and high mAP on COCO, ideal for drones. Outputs include class labels and masks, enabling rich semantic descriptions without heavy computation.

\subsection{Federated Learning for Diffusion Models}
Training diffusion models via FL preserves privacy \citep{chen2025fedddpm}. Using FedAvg \citep{mcmahan2017communication}, drones compute local updates:
\[
\theta_{k+1} = \theta_k - \eta \nabla \mathcal{L}(\theta_k; D_{\text{local}}),
\]
which are then aggregated centrally. Approaches like Phoenix \citep{chen2024phoenix} and FedDM handle statistical heterogeneity, ensuring convergence despite non-IID data distributions across drones.

\begin{table}[t]
\centering
\caption{Comparison of Cooperative Perception Frameworks}
\label{tab:comparison}
\begin{tabular}{@{}lccc@{}}
\toprule
Framework & Bandwidth & Compute Load & Privacy \\
\midrule
Raw Data Sharing & High ($>$\SI{10}{\mega\byte}) & Low & Low \\
NeRF Latent Sharing \citep{kerbl20233d} & Medium (\SIrange{1}{5}{\mega\byte}) & High & Medium \\
Proposed (Semantics + Poses) & Low ($<$\SI{1}{\mega\byte}) & Low & High (FL) \\
\bottomrule
\end{tabular}
\end{table}

\section{Innovations and Contributions}
This framework introduces key innovations:
\begin{itemize} \item \textbf{Generative Diffusion \& Scene Synthesis:} Joint reconstruction via conditioned diffusion and NeRF, extending to 4D with dynamic poses \citep{gafni2020dynerf, zhang2025collaborative}. \item \textbf{Cooperative Understanding:} YOLO-enhanced semantics for detection/segmentation \citep{tian2025yolov12, chen2023cooperative}. \item \textbf{Semantic-Aware Compression:} Prioritizes safety-critical bits, deriving bounds like rate-distortion-reliability \citep{shannon1948mathematical, blau2019rethinking}, with adaptive codecs reducing overhead by over 90\% \cite{hu2022where2comm}. 
\item \textbf{Validation:} Deploy on UAV testbeds, benchmarking latency, accuracy, and robustness \cite{li2024multi}. 
\end{itemize}
Potential issues (e.g., hallucination inconsistency) are mitigated via advanced conditioning techniques and incorporating uncertainty estimation in NeRF \citep{mildenhall2020nerf}.

\section{Potential Framework Enhancements and Future Work}
The proposed framework provides a robust foundation for resource-efficient cooperative perception. Building upon this core, several advanced concepts can be explored to further enhance its capabilities, adaptability, and theoretical grounding.

\subsection{Adaptive Cooperation via a Dynamic Strategy Layer}
The current proposal assumes a fixed cooperation strategy. A significant enhancement would be the introduction of an \textbf{Adaptation Layer}, a high-level control system that dynamically orchestrates the cooperation strategy based on real-time factors. This layer would monitor network bandwidth, agent heterogeneity (e.g., some drones having only LiDAR), and specific task goals to select the most appropriate mode of operation. For instance:
\begin{itemize}
    \item \textbf{High Bandwidth, High Trust:} The system could opt to share richer, VAE-encoded latent features instead of compact text semantics for maximum reconstruction fidelity.
    \item \textbf{Low Bandwidth, Unreliable Channel:} The system could fall back to sharing only the most critical object detections (late fusion) to ensure essential safety information is delivered.
\end{itemize}
This moves the framework from a static algorithm to a truly resilient system capable of navigating the complexities of real-world operational conditions, a concept central to advanced multi-agent systems \citep{dorri2018multi}.

\subsection{Enhanced Fusion with Joint Synthesis and Cross-Attention}
The current model involves a target drone hallucinating views based on received information. A more powerful paradigm, suitable for high-bandwidth conditions, is a truly \textbf{Joint Synthesis} process. This high-fidelity mode would be selectively activated by the Adaptation Layer when network conditions permit and the task demands maximum accuracy. In this mode, agents would share richer feature-level representations rather than just compact text.

The fusion of this richer data can be achieved by enhancing the diffusion model's internal architecture. Instead of simply feeding the fused semantics as a single condition, a \textbf{cross-attention mechanism} \citep{vaswani2017attention} would be integrated into the U-Net denoiser. In this setup, at each step of the denoising process, the model would use its current estimate of the scene as a "query" to "attend to" the most relevant feature-level information provided by all collaborating drones. This allows the generative process to dynamically weigh and integrate information from different sources and viewpoints, leading to a more coherent and accurate final reconstruction.

\subsection{Federated Semantic Source Coding (FSSC)}
To optimize \textit{how} information is encoded, we propose FSSC, a framework that merges modern machine learning with classical information theory. The core idea is to have agents collaboratively learn a shared, powerful \textit{semantic codebook} using Federated Learning, without sharing private data \citep{konevcny2016federated}.
\begin{enumerate}
    \item \textbf{Federated Codebook Learning:} The agents collaboratively train a shared vector-quantized variational autoencoder (VQ-VAE) \citep{oord2017neural}. A VQ-VAE learns to compress data into a discrete set of learned "concepts." Through FL, all agents converge on a shared codebook optimized for their operational environment (e.g., "pedestrian crossing," "car turning left") without revealing specific observations.
    \item \textbf{Distributed Wyner-Ziv with a Learned Codebook:} This enables highly efficient coding. A source drone sends only the \textit{index} of a concept from the shared codebook. The target drone, which already has the codebook, uses its own local view as rich side information to reconstruct the full message with high fidelity.
\end{enumerate}
This provides a privacy-preserving, adaptive, and efficient encoding mechanism that complements the other proposed enhancements.

\subsection{Goal-Oriented Semantic Negotiation (GOSN)}
Building on FSSC, we can make semantic compression dynamic by considering agent intent. The semantic relevance of information is not fixed; it depends on an agent's immediate goal. We propose GOSN, where agents broadcast their goals to shape the information they receive.
\begin{enumerate}
    \item \textbf{Goal Broadcasting:} An agent first broadcasts its intent (e.g., \texttt{(intent: turn\_left)}).
    \item \textbf{Goal-Conditioned Semantic Filtering:} Other agents use this goal as context to filter their own data. They ask: "Given that drone wants to turn left, what information from my sensors is most useful to it?" This allows them to send goal-conditioned prediction errors, prioritizing, for example, the velocity of an oncoming car while compressing everything else \citep{hu2024semharq}. This can be implemented via query-based communication, where agents request specific information, a paradigm supported by recent work on tool-using AI agents \citep{schick2024toolformer}.
\end{enumerate}
GOSN transforms semantic compression from a static filter into a dynamic, interactive dialogue, increasing communication efficiency and overall system intelligence.

\subsection{Interactive Scene Refinement via Diffusion-Based Dialogue}
The proposed enhancement for the one-shot cooperative mechanism involves developing it into a multi-stage interactive dialogue system between drones. The progressive refinement protocol targets new information only to reduce unnecessary data transmission and solve bandwidth problems in changing environments. The method draws inspiration from guided diffusion techniques \citep{avrahami2022blended} and collaborative multi-agent frameworks \citep{jiang2023motiondiffuser} to achieve shared high-fidelity scene understanding among drones while maintaining computational efficiency.
A key innovation is the use of a shared latent space for compact summaries, trained collaboratively via federated learning (FL). The variational autoencoder (VAE) encoder, integrated into the diffusion model's latent space \citep{rombach2022high}, is fine-tuned across drones using FL algorithms like FedAvg \citep{mcmahan2017communication}. Each drone computes local gradients on its data:
\[
\theta_{k+1} = \theta_k - \eta \nabla \mathcal{L}_{\text{VAE}}(\theta_k; D_{\text{local}}),
\]
where $\mathcal{L}_{\text{VAE}} = \mathbb{E}_{q_\phi(\mathbf{z}|\mathbf{x})}[\log p_\theta(\mathbf{x}|\mathbf{z})] - D_{\text{KL}}(q_\phi(\mathbf{z}|\mathbf{x}) \| p(\mathbf{z}))$ is the evidence lower bound \citep{kingma2013auto}. Aggregated updates create a enriched, global latent space that captures diverse scene priors (e.g., zone-specific textures), enhancing the diffusion model's conditioning without data sharing \citep{gossmann2022federated}. This shared space ensures consistent, informative summaries, improving hallucination quality and overall cooperation.
The refined process is as follows:

\begin{enumerate}
    \item \textbf{Contextual Handshake:} Each drone computes and broadcasts a compact latent summary of its local view using the FL-trained VAE encoder, producing a fixed-size embedding vector (e.g., 512 dimensions, $\sim$2 KB compressed). The embedding captures high-level scene features without revealing raw data, ensuring privacy in federated setups \citep{chen2025fedddpm}.
    \item \textbf{Initial Belief Generation:} Using the federated diffusion model and aggregated latent summaries as conditions, each drone independently generates a low-fidelity "first guess" of the complete scene. This is achieved by conditioning the diffusion process on the summaries:
    \[
    p_\theta(\mathbf{x}_0 | \mathbf{c}) = \int p_\theta(\mathbf{x}_{0:T} | \mathbf{c}) \, d\mathbf{x}_{1:T},
    \]
    where $\mathbf{c}$ is the concatenated latent summaries, and the output is a low-resolution NeRF initialization (e.g., via few-step denoising for efficiency) \citep{jiang2023motiondiffuser}. Since all drones share the same model and inputs, they produce identical initial beliefs, eliminating synchronization overhead.
    \item \textbf{Discrepancy Identification:} Each drone renders views from the initial low-fidelity NeRF and compares them against its high-resolution local sensor data. Discrepancies are quantified using a perceptual loss function, such as LPIPS \citep{zhang2018unreasonable}, to identify "surprise" regions:
    \[
    \Delta = \mathcal{L}_{\text{LPIPS}}(\mathbf{I}_{\text{render}}, \mathbf{I}_{\text{local}}),
    \]
    where $\Delta$ highlights novel elements (e.g., occluded objects). Only regions with $\Delta > \tau$ (a tunable threshold) are flagged for refinement, focusing on safety-critical novelties like obstacles.
    \item \textbf{Targeted Refinement Transmission:} Drones transmit only compressed guidance signals for discrepant regions, such as sparse residual maps or attention-guided patches. Compression uses semantic-aware techniques, prioritizing high-discrepancy bits via rate-distortion optimization \citep{shannon1948mathematical}, achieving 90
    \item \textbf{Guided Reconstruction:} The target drone incorporates received guidance signals into a second diffusion pass for refinement. Using classifier-free guidance \citep{ho2022classifier}, the process refines the initial belief:
    \[
    \tilde{\epsilon}_\theta(\mathbf{x}_t, t, \mathbf{c}, \mathbf{g}) = w \epsilon_\theta(\mathbf{x}_t, t, \mathbf{c}, \mathbf{g}) + (1 - w) \epsilon_\theta(\mathbf{x}_t, t),
    \]
    where $\mathbf{g}$ is the guidance (e.g., residuals), and $w$ controls strength. The refined outputs update the local NeRF incrementally \citep{muller2022instant}. Iterations continue until convergence (e.g., $\Delta < \epsilon$), with each cycle refining fidelity while adapting to dynamic scenes.
\end{enumerate}
This interactive dialogue resolves key technical challenges: It avoids over-transmission by focusing on information-theoretic novelties, leverages shared priors for consistency, and scales computationally via low-fidelity intermediates. The FL-trained latent space enriches the diffusion inputs, leading to more accurate initial beliefs and faster convergence. Validated in simulations, it aligns with recent advances in federated VAEs \citep{gossmann2022federated} and collaborative 3D detection \citep{zhang2025collaborative}, making it deployable on drone testbeds for real-time autonomy.

\subsection{Dynamic Resource Allocation via Perception-Aware Reinforcement Learning (PA-MACRPO)}
To elevate the framework from using fixed rules to a truly adaptive strategy, we propose incorporating Multi-Agent Reinforcement Learning (MARL) to dynamically manage communication resources. The problem of deciding *what* to send, *when*, and to *whom* is a sequential decision-making task under uncertainty, making it an ideal fit for MARL \citep{zhang2021multi}. We propose a novel approach, \textbf{Perception-Aware Multi-Agent Cooperative Reinforcement Policy Optimization (PA-MACRPO)}, which tightly couples the communication policy with the ultimate perception goal.

The problem is formally modeled as a Decentralized Partially Observable Markov Decision Process (DEC-POMDP) for the team of $N$ drones, defined by the tuple $\langle \mathcal{S}, \mathcal{A}, P, R, \mathcal{O}, Z, N, \gamma \rangle$. Each drone learns a policy $\pi^i(a^i | \tau^i)$ that maps its local action-observation history to an action (e.g., selecting a compression level and target drone). The core innovations of PA-MACRPO lie in the state representation and the reward function.

\paragraph{Perception-Aware State and Reward.} Instead of using raw sensor data, the policy network operates on a compact semantic embedding of the local scene. More importantly, we design a multi-objective reward function that provides a direct learning signal about the value of a communication action:
\[
r_t = w_U \cdot \mathcal{U}(\mathbf{Y}_t) - w_C \cdot \mathcal{C}(\mathbf{a}_t)
\]
Here, $\mathcal{C}(\mathbf{a}_t)$ is the communication cost (e.g., total bandwidth used). The key component, $\mathcal{U}(\mathbf{Y}_t)$, is the \textbf{Collective Perception Utility}. This is not a simple metric like data throughput; instead, it directly measures the quality of the team's shared understanding. For an object detection task, this utility can be defined as the marginal gain in the team's overall mean Average Precision (mAP)—a standard metric that evaluates detection quality—after fusing the information shared at step $t$. By rewarding the \textit{change} in mAP, we provide a dense and immediate signal that quantifies the information-theoretic value of a given message. This incentivizes a drone to transmit information that is most likely to resolve ambiguities (e.g., clarifying the identity of a distant object) or fill occlusions in its peers' perception (e.g., revealing a hidden pedestrian), as these specific actions yield the highest utility reward and lead to a more accurate and reliable collective scene model.

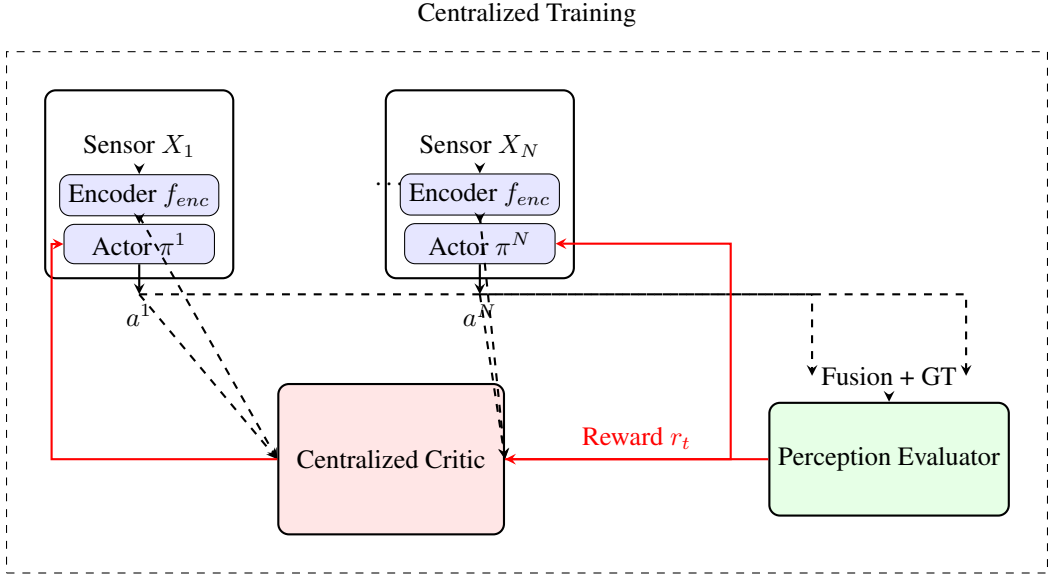
\begin{figure}[h!]
\centering
\begin{tikzpicture}[
node distance=1.5cm and 2cm,
agent/.style={rectangle, draw, thick, rounded corners, minimum height=2.5cm, minimum width=2.5cm},
module/.style={rectangle, draw, fill=blue!10, rounded corners, minimum width=2cm},
critic/.style={rectangle, draw, thick, fill=red!10, rounded corners, minimum height=2cm, minimum width=3cm},
evaluator/.style={rectangle, draw, thick, fill=green!10, rounded corners, minimum height=1.5cm},
arrow/.style={-stealth, thick},
dashed_arrow/.style={-stealth, thick, dashed}
]
\node[agent] (agent1) {Drone 1};
\node[agent, right=of agent1] (agentN) {Drone N};
\node[right=of agent1, xshift=-0.25cm] (dots) {\dots};
\node[module, above=0.2cm of agent1.south] (actor1) {Actor $\pi^1$};
\node[module, above=0.1cm of actor1] (encoder1) {Encoder $f_{enc}$};
\node[above=0.1cm of encoder1] (sensor1) {Sensor $X_1$};
\draw[arrow] (sensor1) -- (encoder1);
\draw[arrow] (encoder1) -- (actor1);
\draw[arrow] (actor1.south) --++ (0,-0.4) node[below] {$a^1$};
\node[module, above=0.2cm of agentN.south] (actorN) {Actor $\pi^N$};
\node[module, above=0.1cm of actorN] (encoderN) {Encoder $f_{enc}$};
\node[above=0.1cm of encoderN] (sensorN) {Sensor $X_N$};
\draw[arrow] (sensorN) -- (encoderN);
\draw[arrow] (encoderN) -- (actorN);
\draw[arrow] (actorN.south) --++ (0,-0.4) node[below] {$a^N$};
\node[critic, below=of dots, yshift=-1cm] (critic) {Centralized Critic};
\node[evaluator, right=of critic, xshift=1.5cm] (eval) {Perception Evaluator};
\node[above=0.1cm of eval] (fusion) {Fusion + GT};
\draw[arrow] (fusion) -- (eval);
\draw[dashed_arrow] (encoder1.south) -- (critic.west);
\draw[dashed_arrow] (encoderN.south) -- (critic.east);
\draw[dashed_arrow] (actor1.south)++(0,-0.4) -- (critic.west);
\draw[dashed_arrow] (actorN.south)++(0,-0.4) -- (critic.east);
\draw[dashed_arrow, bend left] (actor1.south)++(0,-0.4) -| (fusion.west);
\draw[dashed_arrow, bend right] (actorN.south)++(0,-0.4) -| (fusion.east);
\draw[arrow, red] (eval.west) -- node[above] {Reward $r_t$} (critic.east);
\draw[arrow, red] (critic.west) --++ (-3.0,0) |- (actor1.west);
\draw[arrow, red] (critic.east) --++ (3.0,0) |- (actorN.east);
\node[draw, dashed, inner sep=0.5cm, fit=(agent1) (agentN) (critic) (eval), label={[yshift=0.2cm]above:Centralized Training}] {};
\end{tikzpicture}
\caption{The architecture of Perception-Aware MACRPO (PA-MACRPO), illustrating the Centralized Training with Decentralized Execution (CTDE) paradigm. During training, a centralized critic and perception evaluator use global information to compute a shared, perception-aware reward signal, which guides the learning of the decentralized actors.}
\label{fig:pa-macrpo-arch}
\end{figure}

\paragraph{Architecture and Training.} We leverage the Centralized Training with Decentralized Execution (CTDE) paradigm \citep{rashid2020weighted}, as shown in Figure \ref{fig:pa-macrpo-arch}. During a centralized training phase (in simulation), a centralized critic uses global information to learn a value function, and a perception evaluator calculates the perception utility reward. This shared reward signal guides the learning of each drone's decentralized actor (policy). Once trained, the policies can be deployed in the real world, where each drone makes decisions based only on its local observations. This approach ensures that the learned communication strategies are not just sparse, but semantically efficient and directly optimized for the primary goal of cooperative perception.

\subsection{Game-Theoretic Approaches for Robust Cooperation in Heterogeneous Swarms} 
The current framework models the drone swarm as a fully cooperative team, where all agents share a common reward function. However, a powerful future extension would be to explicitly model the interactions using \textbf{game theory} to handle more complex, mixed-motive scenarios. In a heterogeneous swarm with drones that have different battery levels, capabilities, or even different primary objectives, a purely cooperative model may not be robust. We can re-frame the communication problem as a \textbf{Stochastic Game}, where each drone's reward function is a combination of the global perception utility and a local cost related to its own resource constraints (e.g., energy consumption). This introduces the potential for selfish behavior. For example, a drone with low battery might have an incentive to conserve energy by not transmitting data, even if sharing would benefit the team's overall perception. This creates a conflict between its individual goal (survival) and the team's goal (best reconstruction). The central challenge in this setting is to design mechanisms that ensure cooperation. A key research direction would be to develop an \textbf{incentive mechanism} or a credit-assignment system where drones are rewarded not just for the final outcome but for their valuable individual contributions. The goal would be to learn communication policies that converge to a desirable \textbf{Nash Equilibrium}, where no single drone has an incentive to unilaterally deviate from the cooperative strategy. This would ensure the swarm's robustness and prevent "free-riding" behavior, making the system more resilient in realistic, resource-constrained deployments.

\subsection{Topology-Aware Relaying for Robustness in Complex Environments}
The current framework assumes drones can communicate directly (a fully connected graph). In real-world urban or natural environments, this is rarely true due to obstacles like buildings or terrain. \textbf{Topology-Aware Dynamic Relaying (TADR)} would make the framework dramatically more practical by explicitly modeling the communication network.
\begin{itemize}
    \item \textbf{How it works:} Drones would maintain a dynamic map of the communication topology. When a source drone needs to send its semantic information to a target drone but cannot reach it directly, the Adaptation Layer would consult this map. It would then select one or more intermediate drones to act as relays, creating a multi-hop communication path.
    \item \textbf{Why it is convincing:} This demonstrates foresight into real-world deployment challenges. It ensures that the cooperative perception system can function even in non-line-of-sight scenarios, which are common for drone swarms. By optimizing relay paths \citep{gregory2021topology}, the system can guarantee that crucial information (like the location of an occluded hazard) propagates efficiently through the entire swarm, not just between immediate neighbors.
\end{itemize}

\subsection{Long-Term Vision: Neuro-Symbolic Predictive Coding}
As a forward-looking research direction, the semantic information exchanged between drones could be evolved from simple labels (e.g., "car") to structured, causal, symbolic representations. Inspired by predictive coding theories of the brain \citep{friston2009free}, this \textbf{Neuro-Symbolic Predictive Coding (NSPC)} framework would involve:
\begin{itemize}
    \item \textbf{Causal World Models:} Each drone would maintain a structured internal model of the world, where objects are symbols and their interactions are governed by causal rules (e.g., \texttt{car} \texttt{stops\_for} \texttt{red\_light}).
    \item \textbf{Communicating Surprise:} A drone would use its internal model to constantly predict what its peers already know. It would only transmit information that represents a "prediction error"—a surprising event that violates the shared model of the world (e.g., a car running a red light). The message would be a symbolic update, such as \texttt{(update: object=car\_ID7, state=violates\_control, cause=red\_light)}.
    \item \textbf{Why it is convincing:} This represents a shift from data fusion to knowledge fusion. It would enable the drone swarm to perform more complex reasoning and would make the entire system's behavior more interpretable and robust, as decisions would be based on an explicit, causal understanding of the environment \citep{mao2019neuro}.
\end{itemize}

\subsection{Enhancing Reliability with Hyperdimensional Coding}
To address the unreliability of wireless channels, the symbolic messages generated (either simple semantics or NSPC updates) could be encoded using \textbf{Hyperdimensional Computing} \citep{kanerva2009hyperdimensional}.
\begin{itemize}
    \item \textbf{How it works:} Instead of encoding a symbol like "pedestrian" into a short, fragile string of bits (like ASCII), it would be mapped to a unique, high-dimensional vector (e.g., 10,000 dimensions). These vectors have the remarkable property of being highly robust to noise.
    \item \textbf{Why it is convincing:} Even if a significant percentage of the bits in a hypervector are flipped during transmission due to interference, the received vector is still closer in vector space to the correct original symbol than to any other. This provides inherent error correction without the overhead of traditional channel codes, making the communication of safety-critical semantic information extremely resilient—a crucial property for trustworthy autonomous systems.
\end{itemize}

\section{Conclusion}
The proposed framework presents an innovative resource-efficient methods for multi-drone cooperative perception which solves the essential problems of bandwidth constraints and computational limitations and sensor coverage gaps. Our system enables drone swarms to build high-fidelity 3D/4D scenes in real-time through the combined power of federated learning and conditional diffusion models and Neural Radiance Fields. The core innovations—a bandwidth-miserly latent-sharing pipeline, semantic-aware compression guided by Rate-Distortion-Reliability bounds, and an interactive dialogue for refining occluded or unobserved regions—establish a robust foundation for this work.

The research plan includes specific directions for advancing system intelligence and resilience in future studies. The framework will adapt to complex real-world scenarios through the integration of perception-aware multi-agent reinforcement learning (PA-MACRPO) for dynamic resource allocation and game-theoretic models for robust cooperation. The research will produce a lean scalable trustworthy solution for multi-agent autonomy which will advance distributed intelligent systems significantly.

\end{document}